\documentclass[conference]{IEEEtran}
\IEEEoverridecommandlockouts
\usepackage{cite}
\usepackage{amsmath,amssymb,amsfonts}
\usepackage{algorithmic}
\usepackage{graphicx}
\usepackage{textcomp}
\usepackage{xcolor}
\usepackage{caption}
\usepackage{subcaption}
\usepackage{color}
\usepackage{soul}
\def\wei{\textcolor{red}}
\def\cc{\textcolor{black}}
\def\BibTeX{{\rm B\kern-.05em{\sc i\kern-.025em b}\kern-.08em
    T\kern-.1667em\lower.7ex\hbox{E}\kern-.125emX}}

\usepackage[bottom]{footmisc}

\usepackage{tabularx}
\usepackage{booktabs}
\usepackage{ragged2e}
\usepackage{threeparttable}

\newcommand{\comment}[1]{}
\newcolumntype{L}{>{\RaggedRight\hangafter=1\hangindent=0em}X}
\def \fulltablewidth {17.2cm} 

\def \leftcolumnl {0.2cm}   

\newcommand{\NR}{$G$}    
\newcommand{\WR}{$G{'}$}    
\newcommand{\Rel}{$K$}    
\newcommand{\Gold}{$R$}    

\newcommand{\colorone}{0.976,0.506,0.443}
\newcommand{\colortwo}{0.596,0.745,0.298}
\newcommand{\colorthr}{0.4,0.588,0.91}

\begin{document}

\title{The Exploration of Knowledge-Preserving Prompts for Document Summarisation}


\author{\IEEEauthorblockN{ Chen Chen, Wei Emma Zhang, Alireza Seyed Shakeri}
\IEEEauthorblockA{\textit{School of Computer and Mathematical Sciences} \\
\textit{The University of Adelaide}, Adelaide, Australia \\
c.chen.adelaide@gmail.com;wei.e.zhang@adelaide.edu.au \\
alireza.seyedshakeri@adelaide.edu.au}
\and
\IEEEauthorblockN{Makhmoor Fiza}
\IEEEauthorblockA{\textit{Department of Management Sciences and Technology} \\
\textit{Begum Nusrat Bhutto Women University}\\
Sukkur, Pakistan \\
makhmoor.fiza@bnbwu.edu.pk}
}

\maketitle

\begin{abstract}
Despite the great development of document summarisation techniques nowadays, factual
inconsistencies between the generated summaries and the original texts still occur from time to time. This study explores the possibility of adopting prompts to incorporate factual knowledge into generated summaries. 
We specifically study prefix-tuning that uses a set of trainable continuous prefix prompts together with discrete natural language prompts to aid summary generation. 
Experimental results demonstrate that the trainable prefixes can help the summarisation model extract information from  discrete prompts precisely, thus generating knowledge-preserving summaries that are factually consistent with the discrete prompts. 
The ROUGE improvements of the generated summaries indicate that explicitly adding factual knowledge into the summarisation process could boost the overall performance,   showing great potential for applying it to other natural language processing tasks.
\end{abstract}


\section{Introduction}
Document summarisation is a technique that can help people quickly browse and filter information by summarizing and extracting the key content of articles, and is therefore a useful technique for today's explosive growth of information on the Internet. There are two main document summarisation approaches, extractive summarisation and abstractive  summarisation~\cite{YutongQuv2}. 
\comment{
Decades ago, there were more studies on extractive abstraction. In recent years, benefiting from the proposal of Transformer~\cite{TransformerX} and the emergence of various large-scale pre-trained language models such as GPT~\cite{GPT}, BART~\cite{BART}, more research now focuses on using pre-trained language models to do abstractive summarisation.}
\cc{Extractive summarisation is faithful to the original text, using words or phrases extracted from the source to form a summary, while abstractive summarisation summarizes the semantics of the original text, without being limited by vocabulary~\cite{YutongQuv2}.}

Abstractive summarisation allows us to generate natural summaries that are more like human-written ones, rather than extracting information from the source as extractive summarisation does. However, one problem with abstractive summarisation is that the generated summaries are not always factually consistent~\cite{Enhancing}. 
To solve this problem, one approach is to incorporate knowledge into the model to aid summary generation. For example, KG-BART~\cite{KG-BART} incorporates knowledge by adding a set of KG-encoder and KG-decoder to BART to generate more logical and common sense sentences, while~\cite{Enhancing}  designs a graph attention network (GAT) to take external knowledge and enhance the factual consistency of the generated summaries by passing the output of the GAT to the cross attention layer of a decoder. However, in the above studies, adding knowledge to a model usually requires not only embedding the knowledge but also designing new structures to incorporate it. 

Stopping to rethink this sequence-to-sequence problem of document summarization, we realise that when generating summaries using pre-trained language models, we have been trying to figure out how to generate correct summaries using the documents together with the model's prior (a pre-trained language model is usually trained on large corpus and therefore contains a large amount of prior knowledge in the model \cite{CoCo}). If the model's prior is correct or the input documents  could effectively influence the summary generation, then the generated summaries would be correct.
and conversely, the summaries would have inconsistencies with the original text~\cite{CoCo}. Thus, the problem of adding knowledge to a model can be transformed into the problem of how to combine texts more effectively with the model prior. Following this idea, we propose a knowledge-enhanced document summarization solution based on prefix-tuning, which is a  type of prompt-based learning~\cite{PromptSurveyX}, and open information extraction, which extracts knowledge from sentences. 

Prompting methods 
 leverage the text generation capability of pre-trained language models to design appropriate prompts for downstream applications~\cite{PromptSurveyX}. 
There are continuous prompts and discrete prompts. 
{Discrete prompts are natural language prompts composed of words, {and continuous prompts could be considered as vectors that serve as prompts.}}
Prefix-tuning is to train a set of continuous task-specific prefix prompts to steer the model to generate texts for different generation tasks~\cite{Li&Liang}. 
To incorporate knowledge into summary generation, we simply add 
discrete prompts that represent the factual knowledge and are extracted by OpenIE~\cite{OpenIE} in front of the original input text. 
The purpose is to explore whether explicitly prepending factual prompts could help the model generate a summary that is factually consistent with the prepended prompts. %

To test the effectiveness of this approach, we use two metrics, namely CoCo~\cite{CoCo} and ROUGE~\cite{ROUGE} to evaluate the generated summaries. CoCo scores the factual consistency of the generated summaries by averaging the positional scores of \cc{selected} important words \cc{such as nouns or verbs that are crucial to the factual consistency of the generated summaries}, while ROUGE is based on the overlap of n-grams. Experiments demonstrate that since the calculation of the positional scores relies on the occurrence of important words in the original text, the scoring model will prefer summaries generated by other models that contain more words from the original text, and instead considers the more abstract golden summaries written by humans to have lower factual consistency. But the ROUGE scores of the summaries generated with added relation are indisputably higher than when there is no relation added.

This work has contributions in the following aspects:
\begin{itemize}
    \item We propose a simple way to incorporate factual knowledge into document summarisation and largely improve the model performance;
    \item We extensively explore the possibility and performance of adapting prefix-tuning and prepending explicit knowledge for knowledge preservence;
    \item We provide analysis studies to demonstrate the CoCo could serve as a factual consistency metric in some cases, but could not work for all. Our studies fill the gap of relevant studies on document summarization applications.  
\end{itemize}

\section{Related Work}
Our research is mainly related to two directions: prompt-based learning and controllable text generation.

\subsection{Prompt-based learning}

The prompting method is a new paradigm for using pre-trained language models and was summarized and presented by~\cite{PromptSurveyX}. The philosophy of prompt-based learning is to modify tasks to fit models instead of modifying models to fit tasks. \cc{For example, we can design cloze prompts and have a model perform a ‘fill-in-the-blank’ task to output answers itself~\cite{LAMA}, or alternatively add prefix prompts before a model to steer the model to do text generations~\cite{Li&Liang}. The prompts added to a model can be not only discrete natural language prompts~\cite{LAMA}, but also continuous vectors~\cite{Li&Liang}, or a mixture of discrete and continuous prompts~\cite{P-tuningX}.} To find the best prompt for a specific task, we can either manually design different prompts and try out their performance on the task~\cite{LAMA}, or we can use automated methods such as prompt mining~\cite{PromptMining} or prompt tuning~\cite{Li&Liang} to continuously optimize a prompt. The  prompt-based learning has been widely adopted in more than 20 natural language processing-related tasks including text classification, knowledge probing, and so forth with good results~\cite{PromptSurveyX}. In our study, we focus on prefix-tuning, combining trainable continuous prefixes with discrete natural language prefixes to do controllable generation.

\subsection{Controllable text generation }

Controllable generation is a topic that has been studied extensively. The prefix-tuning proposed in recent years has opened up more possibilities in this field. By training a set of contrastive prefixes,~\cite{ContrastivePrefixes} improved the controllability of the prefixes and guided the model to generate texts of specific sentiments and attributes.~\cite{ControlPrefixes} then customize prefixes for different documents, and achieve controllable generation by combining a task-specific prefix with particularly controlled prefixes. A study similar to~\cite{ControlPrefixes} is Tailor~\cite{Tailor}, which steers the model to generate sentences with multiple attributes by ‘weaving’ together the prefix prompts of different attributes. However, most of the related studies have all worked on changing prefixes, and few studies have looked into the properties of prefix-tuning and what happens if we directly change the document, which is what we explore in this paper. 

\begin{figure*}
\centering
\includegraphics[width=0.8\textwidth]{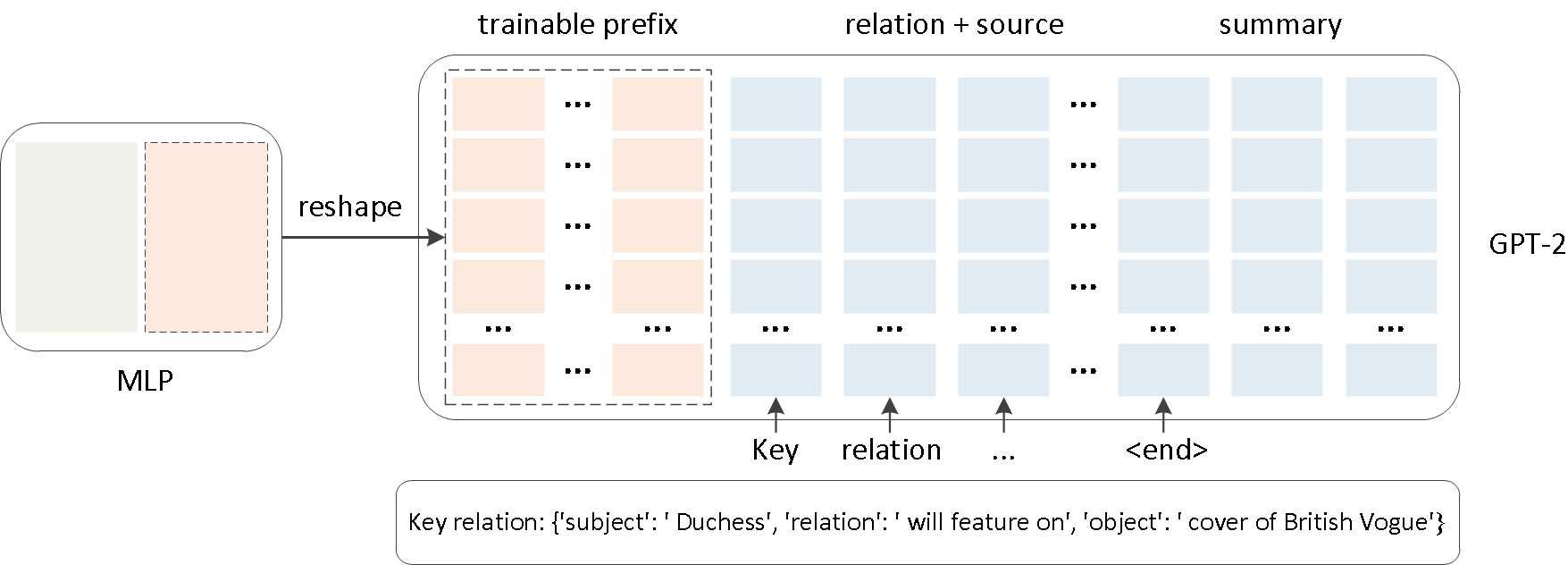}
\caption{Adding knowledge via prefix-tuning. The salmon pink dash-squared block in MLP represents the last layer of the MLP. Prefix-tuning reshapes it and passes it to GPT-2. The short text below is an example of a relation extracted from a sentence.}  
\centering\label{flow}
\end{figure*}

\section{Methodology}
\cc{In our study, we use GPT-2 as the base model for document summarisation. }Figure~\ref{flow} depicts the  process of our method to add factual knowledge into \cc{GPT-2}. The factual knowledge is obtained by adopting the technique of open information extraction (open IE) and output in the form of triplets containing entities we care about. 
After adding the relations to the
texts, we train a set of continuous prefix prompts
on the modified text. We evaluate whether this set of prefixes
could recognize the information we add \cc{to the texts} and could help
the model to generate summaries that are factually
consistent with the added factual knowledge. 


\subsection{Extracting factual knowledge}

In this paper, we use OpenIE \cite{emnlp/KolluruAAMC20} to extract the relations
that exist in the sentences, and filter relations by keeping only interested types of entities. 

Open information extraction (open IE) refers to the extraction of relation tuples, typically binary relations, from plain text, such as  (Barack Obama; was born in; Hawaii) where ``Barack Obama" and ``Hawaii" are subject and object respectively while ``was born in" is the relation. 
The central difference from supervised information extraction is that the schema for these relations does not need to be specified in advance. Instead, the relation is typically the text linking two arguments (i.e., subject and object).  Recall that for supervised relation extraction, there are a fixed small number of relation types that serve as the classification labels. 

Open IE techniques usually rely on lexical, syntactic or linguistic information which is regarded in general as the rule-based methods as no training signals are required. 
There is a large body of work on open information extraction. TextRunner \cite{naacl/YatesBBCES07} and ReVerb \cite{emnlp/FaderSE11} which make use of computationally efficient surface patterns over tokens. 
 Ollie \cite{emnlp/MausamSSBE12} was built upon fast dependency parsers. 
However, these approaches are brittle on the out-of-domain text and long-range dependencies as they require a large set of patterns (i.e. rules). 
 OpenIE \cite{OpenIE} 
 thus  addressing the issue by replacing this large pattern set with a few patterns for canonically structured sentences.
 Specifically, the method first splits each sentence into a set of entailed clauses. This step is formed as searching certain arcs in the dependency trees. To get the correct patterns, the authors defined eight pattern classes and trained a classifier on the existing labelled relation extraction dataset -- this could be considered as a training step on a stand-out dataset. 
 Then each clause is maximally shortened by adopting natural logic \cite{phdvalencia}, producing a set of entailed shorter sentence fragments. These fragments are then segmented into OpenIE triples by utilizing a small set of pattern rules with 14 patterns. 

 After obtaining relations, we filter the multiple extracted relations by applying named entity recognition (NER). The motivation is that the relations which contain important entities could be  important knowledge to be kept. 
 For example, the NER tag ‘PERSON’ could identify the main subject of a sentence.  


 \begin{figure*}[t]
  \centering
  \begin{subfigure}{\linewidth}    
      \centering
    \includegraphics[width = .85\linewidth]{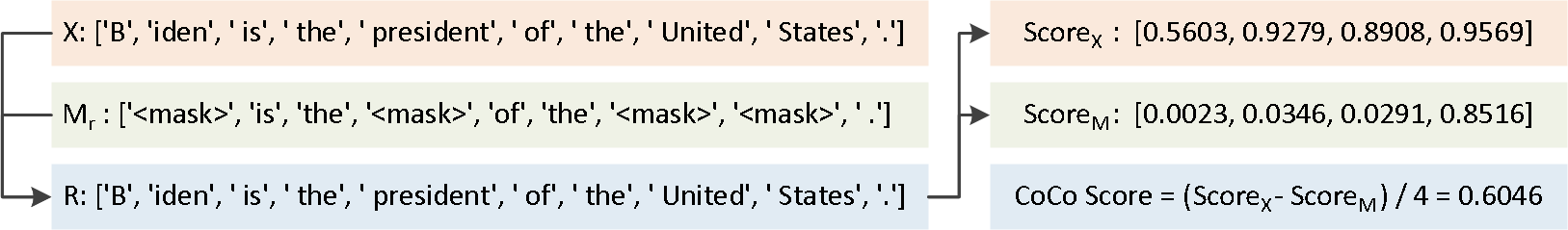}
    \caption{CoCo score for the reference. $Score_{X}$ and $Score_{M}$ are the positional scores of `B', `president', `United', `States' on $X$ and $M_{r}$}
    \label{fig:CoCoa}
  \end{subfigure}
  \begin{subfigure}{\linewidth}
      \centering
    \includegraphics[width = .85\linewidth]{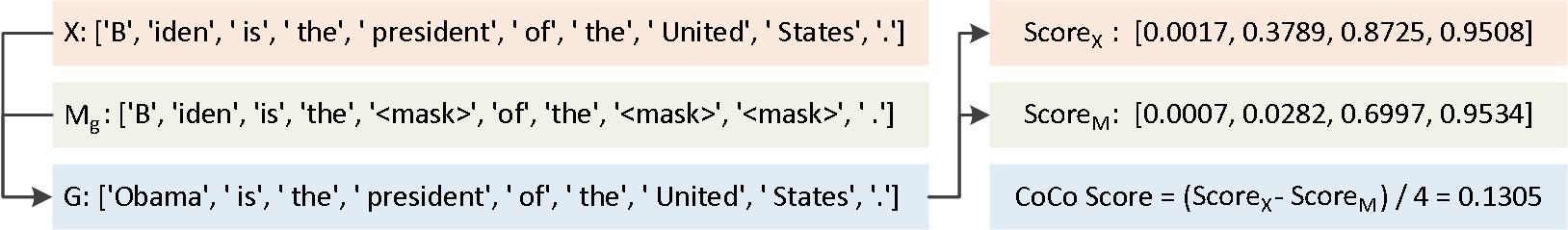}
    \caption{CoCo score for the summary. $Score_{X}$ and $Score_{M}$ are the positional scores of `Obama', `president', `United', `States' on $X$ and $M_{g}$}
    \label{fig:CoCob}
  \end{subfigure}
  \caption{A simple example of CoCo calculation. $X$ refers to the original text. $M$ denotes the masked text. $R$ and $G$ are reference summaries and model-generated summaries respectively.}
  \label{fig:CoCo}
\end{figure*}  

\subsection{Applying prefix-tuning}
 Prefix-tuning trains task-specific prefixes to perform different generation tasks. For document summarisation tasks, generally, 
 {we can use ‘TL;DR:' as a text prompt to perform zero-shot document summarisation when using GPT-2~\cite{GPT-2}.}
Different from adding text or discrete prompts, prefix-tuning uses a set of trainable parameters as a prefix (i.e., continuous prompts) in front of the text. 

During training, following the prefix-tuning paper~\cite{PromptSurveyX}, we keep \cc{most of} the parameters of \cc{GPT-2} constant and train only the parameters in the prefix. We then obtain a set of prefixes that allow \cc{GPT-2} to \cc{perform document summarisation task} as normal. Specifically, since the structure of GPT-2 is a multi-layer Transformer decoder, its prediction of the next word is jointly determined by all the preceding words, so as long as we can make targeted changes to the left context, then we can control the generation of the model. ‘past\_key\_values’ of GPT-2 was originally used to store the previous computation results of the model to speed up computation, but prefix-tuning cleverly exploits this by mapping a set of vectors to the shape required by ‘past\_key\_values’ through a fully connected neural network MLP and passing the vectors to the model to achieve the purpose of adding prefix before the input text. 
Having been passed to the model, these vectors will be concatenated with the existing keys and values of the model in the dimension of sequence length, thus controlling the model generation.

\subsection{Incorporating factual knowledge}
Although language models contain a large amount of knowledge, this knowledge is also the reason why the summaries generated by the pre-trained language models do not always agree with the facts of the original text~\cite{CoCo}. Therefore, how to use the knowledge we need and avoid adding wrong priors becomes a problem.

To do this, we emphasize factual knowledge  by adding a set of natural language prompts before the texts.  As shown in  Figure~\ref{flow}, we explicitly add the relation (starting as `Key relation:') to inform the model to generate a factual consistent summary by keeping the knowledge represented in the prepended relation.  

\subsection{Adapting metrics}

A common metric used in natural language generation is ROUGE~\cite{ROUGE}, which scores the generated summaries based on the overlap of n-grams. ROUGE does not measure the factual consistency between golden summaries and model-generated summaries.  So we include CoCo~\cite{CoCo}, which is able to evaluate the factual consistency. 

CoCo performs the evaluation based on the positional scores of important words. The positional score gives the logarithmic probability of each word in a summary occurring at a particular location given the original text $X$. The basic idea of CoCo is to first detect the keywords in the summary, mask them in the original text to generate a masked document $M$, and then use $M$ and the original text $X$ to calculate the difference of positional scores of important words in the {generated summaries $G$}. Part of speech tagging (POS) and NER are used to pick the important words to be masked.
If the positional score of a word appearing at a certain position in the summary under $X$ is significantly higher than the positional score of the word appearing at that position under $M$, then CoCo assumes that the word at that position in the summary is more influenced by the original text, the fact related to this word is more likely consistent with the fact of the original text ~\cite{CoCo}. 
%
{To calculate the positional score of the words in $G$ with respect to $X$ or $M$, we need an independent model to score the summary $G$ on $X$ or $M$ \cite{CoCo}. In our study, we use BART, as BART is trained specifically for summarization tasks and there are versions of the model available that have been fine-tuned on CNN/Daily Mail and XSum}. 
%

Figure~\ref{fig:CoCo} shows an example of CoCo calculation. 
Assume the original text $X$ is `Biden is the president of the United States.', the golden summary $R$ happens to be exactly the same as the original text\footnote{We use this simplified example to describe how CoCo is calculated. In a real dataset, the original text is much longer than the summary.}. The generated text $G$ is `Obama is the president of the United States.', then CoCo will first use POS or NER tags to pick out the important words. Suppose we use NER and keep all the ‘PERSON’, ‘TITLE’ and ‘COUNTRY’ in the results, then the keywords picked from $R$ will be ‘Biden’, ‘president’, ‘United’, ‘States’, and the keywords picked from $G$ will be ‘Obama’, ‘president’, ‘United’, ‘States’. 
 Masking off the corresponding words in $X$, we get $M_{r}$: `$\textless mask\textgreater$ is the $\textless mask\textgreater$ of the $\textless mask\textgreater$ $\textless mask\textgreater$.' and $M_{g}$: `Biden is the $\textless mask\textgreater$ of the $\textless mask\textgreater$ $\textless mask\textgreater$.' as shown in the figure.  
 
Assuming $N$ keywords in the target text are picked, taking only the first BPE token of each keyword, the CoCo score could be calculated as:
$CoCo = (Score_X- Score_M)/N$. 
Take the calculation of $R$'s CoCo score as an example, tokens of the sentence $R$ after BPE are ['B', 'iden', ' is', ' the', ' president', ' of', ' the', ' United', ' States', '.'] as shown in Figure~\ref{fig:CoCoa}. 
Here we can see that since the word `Biden' appears in $X$, 
the positional score for the first token 'B' given by CoCo is high, while in $M$, the first word is `$\textless mask\textgreater$', and the score for `B' is considerably lower. 
Thus, the larger the difference between $Score_{X}$ and $Score_{M}$, the more likely the sentence is factually consistent. 
On the contrary, for the score of $G$, since the two positional scores for 'Obama' against 'Biden' are equivalently small, the $Score_{X}-Score_{M}$ is small and the CoCo score is low, which means the sentence is likely to be factually inconsistent with the original.

\section{Experiment}
We first introduce the datasets used and the baselines compared in the experiments, then present our examination on the properties of prefix-tuning. 
Later, we report the results of the knowledge-enhanced document summarisation. 
\subsection{Datasets}

\subsubsection{CNN/Daily Mail}
CNN/Daily Mail is a dataset proposed for abstractive summarisation~\cite{CNN/DM}. 
The source documents are the CNN and Daily Mail news crawled from their websites. The summaries are human-generated and formed as bullets. 
The CNN/Daily Mail dataset contains 286,817 training data, 13,368 validation data and 11,487 test data. The average length of the source texts is 766 words, and summaries are usually three to four sentences, with an average length of 53 words~\cite{CNN/DM}. 
Since GPT-2 has a limit on the input length, and training for prefix-tuning does not require a large size of parameters,  our evaluation uses part of the data, i.e., 50,000 data instances that satisfies the sum of the length of the source and target does not exceed 800. If the added relation is very long, then we further reduce the data usage according to the length of the new text in order to satisfy the input length requirement of GPT-2. 

\subsubsection{XSum}
XSum~\cite{XSum} is another commonly used dataset in the field of document summarisation. It is sourced from BBC's online articles~\cite{XSum}. XSum  contains 204,045 training data, 11,332 validation data and 11,334 test data.  XSum seeks to summarize an article using a very short sentence, so its target is shorter than that of CNN/Daily Mail, and also more abstract.  
The average length of XSum's source texts is 431 words, but the average length of its targets is only about 23 words~\cite{XSum}. The abstract nature of XSum makes the abstractive summarisation task more challenging, so getting good results on XSum can also make a model more convincing. Similar to processing CNN/Daily Mail, we only use the first 50,000 data instances which satisfy the sum of the length of the source and the target is less than 800 for training.

\subsection{Baselines}

To evaluate the performance of the model, we use two metrics in our experiments. ROUGE to evaluate the degree of overlap between the generated summaries and the golden summaries, and CoCo to evaluate the factual consistency of the generated summaries with the original texts. Since there are now many models trained specifically for document summarisation, it is difficult for GPT-2 to surpass their performances. Therefore, when comparing ROUGE scores we only compare with the ROUGE scores reported by GPT-2  when using the `TL;DR:' prompt. 

Given that prefix-tuning is far more expressive than  discrete prompt, the ROUGE scores for prefix-tuning should at least exceed the `TL;DR:' prompt. 
If relations are added to the texts,  theoretically the summaries generated from the modified dataset $X{'}$ (where $X{'}$ equals the relation plus original input $X$) should have higher ROUGE scores than the situation when no relations are added.

CoCo is not a commonly used metric like ROUGE. It is rare to see research that uses CoCo to compare the factual consistency of generated summaries in this stage, so for the CoCo score, we use the summaries generated without added relations as the baseline. The CoCo score for the summaries generated with added relations should be higher than when there is no relation added.

\subsection{Preliminary Experiments}

The preliminary experiments are performed on CNN/Daily Mail and have 
 two parts. The first
is abstractive summarisation using prefix-tuning. The second is
sentence extraction. The purpose of evaluating the capability of sentence extraction is to showcase that prefix-tuning is suitable for {knowledge-enhanced document} summarisation as it could identify the key information. 
The experimental results
show that prefix-tuning is effective for document
summarisation task on GPT-2, and it is good at
extracting information from structured content in
texts. These properties of prefix-tuning fit well
with our purpose.

\subsubsection{Prefix-tuning for abstractive summarisation}

Abstractive summarisation is a preliminary experiment to examine the performance of prefix-tuning
for document summarisation task without adding relations. The results can be used as a baseline for knowledge-enhanced document summarisation.

When using prefix-tuning for document summarisation on CNN/Daily Mail, the ROUGE-1 is only about 20 if we only use a prefix of length 5, but if a longer prefix is used, then the performance of the model immediately improves significantly. If we use a prefix of length 100 and allow the model to generate summaries with a maximum length of 100, then the ROUGE-1 for summaries generated on CNN/Daily Mail using GPT-2 can reach 30, which is consistent with what~\cite{Li&Liang} reported that the longer the prefix, the more expressive it is. For the part of the data we use, the ROUGE scores exceed what is reported by GPT-2 for document summarisation on CNN/Daily Mail using `TL;DR:'.  Table~\ref{tab:TL:DR} shows the results.

\begin{table}[t] \small
\begin{center}
\caption{Results of TL;DR: and Prefix-tuning on CNN/Daily Mail. Pre-$x$ means the prefix length is $x$ }
	\label{tab:TL:DR}
\begin{tabular}{lccc} 
 \hline
    & \textbf{ROUGE-1} & \textbf{ROUGE-2} & \textbf{ROUGE-L} \\ [0.5ex] 
 \hline
 TL;DR: & 29.34 & 8.27 & 26.58 \\
		Pre-5 & 21.85 & 9.28 & 20.17 \\
		Pre-100 & 32.89 & 14.58 & 30.78 \\
\hline
\end{tabular}
\end{center}
\end{table}

\begin{table}[t] \small
\begin{center}
\caption{Results of sentence extraction on CNN/Daily Mail$^*$}
	\label{tab:Extractive}
\begin{tabular}{c c c c} 
 \hline
    & \textbf{ROUGE-1} & \textbf{ROUGE-2} & \textbf{ROUGE-L} \\ [0.5ex] 
 \hline
    SenEx1 & 99.98 & 99.79 & 99.98 \\
		SenEx2 & 96.56 & 95.64 & 96.46 \\
		SenEx3 & 65.50 & 59.38 & 64.25 \\
\hline
\end{tabular}
\begin{tablenotes}
       \item [] *The tokens used here are results of byte pair encoding, so they are not guaranteed to be three complete words.
\end{tablenotes}
\end{center}
\end{table}

\subsubsection{Sentence Extraction}

Sentence extraction is to further explore the properties of prefix-tuning. Specifically, this preliminary experiment uses the method proposed in the methodology to test whether prefix-tuning can truly identify the information we want to extract accurately, especially its boundaries. The experiments show that, when experimenting with CNN/Daily Mail for sentence extraction, if the model is trained directly with the first sentence of source texts without any guidance (SenEx1), the model can extract that sentence almost perfectly. If the first three tokens of a sentence are added to the front of the source and then the model is trained (SenEx2), then the model can extract that sentence most of the time, but with reduced accuracy. If any three tokens from a sentence are added to the source and the model is trained with that sentence (SenEx3), the model can still extract that sentence, but the ROUGE-1 score drops to about 65, as Table~\ref{tab:Extractive} shows.

It is worth noting that, first of all, the extraction is exact to the token. For example, the periods ‘.’ and ‘ .’ (with a space) are two tokens in GPT-2 embedding. If we take the first period ‘.’ as the criterion for extracting a sentence, the generated summaries will  definitely be bounded by the first period, and will never stop at the second period ‘ .’. Second, if the model does not correctly extract the sentence we want to extract, the result is still necessarily a sentence from the original text and will not be generated arbitrarily. Therefore, we believe that when using any three tokens of a sentence for extraction, the model is still able to recognize the boundaries, even though in many cases the model does not correctly identify the sentence we are  extracting. As long as we provide more special tokens (e.g., tokens of low-frequency words instead of tokens of conjunctions or punctuation), the model will be able to identify which sentence we are extracting and extract that sentence precisely. 

The above preliminary experiments show that prefix-tuning can indeed extract key information directly from the original text, and its extraction is so precise that it can identify the patterns of structured information in the text, extract the important information with token precision, and filter out the frame that contains this information in the original text. These properties of prefix-tuning could be useful for incorporating knowledge into document summarisation.

\begin{table} [t] \small
\begin{center}
\caption{ROUGE scores}
	\label{tab:ROUGE}
\begin{tabular}{lccc}
\hline
      & \textbf{ROUGE-1}  & \textbf{ROUGE-2}  & \textbf{ROUGE-L}  \\ \hline
\NR   & 25.07    & 12.13    & 23.46    \\
\WR  & 62.38    & 49.46    & 61.44    \\ \hline
\\
\multicolumn{4}{c}{(a) CNN/Daily Mail} \\
\\
\hline
      & \textbf{ROUGE-1}  & \textbf{ROUGE-2}  & \textbf{ROUGE-L}  \\ \hline
\NR   & 23.77    & 7.06     & 19.73    \\
\WR  & 45.69    & 31.06    & 43.01    \\ \hline
\\
\multicolumn{4}{c}{(b) XSum}          
\end{tabular}
\end{center} 
\end{table}

\subsection{Knowledge-enhanced Document summarisation}
We present our findings and analysis of the model performances of two evaluation metrics. ROUGE shows the general performance of the summarisation. CoCo indicates the knowledge-preserving capability.

\subsubsection{ROUGE}
The experimental results in Table~\ref{tab:ROUGE} show that the ROUGE scores of the summaries generated on $X{'}$ ($X$ with added relations) are significantly higher than the scores of the summaries generated on the original document $X$. There could be two reasons: i) After adding structured discrete prompts in front of the original text $X$, the trainable continuous prompts accurately identify the keywords extracted from the targets contained in the discrete prompts. These words appear in the summary $G{'}$ as is, making $G{'}$ a significant improvement in the overlap with targets. 
ii) Although many words in targets are not included in the added relations, these words are related to the added relations. Since $G{'}$ is generated based on the added relations as we will see in the case study, it is likely that these words will be included in the summaries $G{'}$, further increasing the degree of overlap between $G{'}$ and targets. In addition, we can see that the improvement of ROUGE scores on CNN/Daily Mail is larger than that on XSum, which is due to the abstract nature of XSum. The golden summaries of XSum are much more abstract than that of CNN/Daily Mail, therefore sometimes $G{'}$ generated on XSum are similar to the targets but with different  vocabulary.

\subsubsection{CoCo}
Since the relations extracted using OpenIE are subject-predicate-object triads, where the subject and object are mainly nouns and the predicate is mainly a verb, we also primarily compare the ability of the summaries to preserve the factual consistency of nouns and verbs. We first compare whether $G'$ would have a higher CoCo score than $G$ when retaining only the POS tags ‘NOUN’ and ‘PRON’ as intuitively subject and object errors in a sentence are easier to detect, then we retain ‘VERB’ in addition to see if the positional scores of verbs can improve the CoCo score of the summaries. The results are shown in Table~\ref{tab:CoCo}. 
As we can see in the table, $G{'}$ have lower CoCo scores than $G$, and the scores are even lower if the verbs are retained, despite $G{'}$ being closer to the golden summary than $G$. CoCo reports a higher correlation coefficient with human judgment when measuring the factual consistency of a sentence with the original text compared to other criteria~\cite{CoCo}, but the results in our study suggest that CoCo may not be suitable for comparing the factual consistency of two sentences. 
We will further discuss this issue in the section of other discussions.  

\begin{table} [t] \small
\begin{center}
\caption{CoCo scores}
	\label{tab:CoCo}
\begin{tabular}{lcc}
\hline
     & \multicolumn{1}{l}{\textbf{Only Noun}} & \multicolumn{1}{l}{\textbf{Noun+Verb}} \\ \hline
\NR  & 0.0264                                 & 0.0253                                 \\
\WR & 0.0179                                 & 0.0163                                 \\ \hline
\\
\multicolumn{3}{c}{(a) CNN/Daily Mai}  
\\
\\ \hline                    
     & \multicolumn{1}{l}{\textbf{Only Noun}} & \multicolumn{1}{l}{\textbf{Noun+Verb}} \\ \hline
\NR  & 0.0086                                 & 0.0074                                 \\
\WR & 0.0074                                 & 0.0063                                 \\ \hline
\\
\multicolumn{3}{c}{(b) XSum}                                          
\end{tabular}
\end{center}
\end{table}

\begin{table*}[!htbp]
	\centering
	\setlength{\aboverulesep}{2pt}
    \setlength{\belowrulesep}{2pt}
	\begin{subtable}{\linewidth}
    \centering
	\begin{tabularx}{\fulltablewidth}{p{\leftcolumnl} L}  
	    \toprule
        \Rel: &  Key relation: \{`subject': ` \textcolor[rgb]{\colorone}{Sally Forrest}', `relation': ` \textcolor[rgb]{\colortwo}{died on}', `object': ` \textcolor[rgb]{\colorthr}{March 15}'\}  \\
        \midrule
        \Gold: &  \textcolor[rgb]{\colorone}{Sally Forrest}, an actress-dancer who graced the silver screen throughout the '40s and '50s in MGM musicals and films \textcolor[rgb]{\colortwo}{died on} \textcolor[rgb]{\colorthr}{March 15}. \\
        \midrule
        \NR: &  Actress: \textcolor[rgb]{\colorone}{Sally Forrest} was in the 1951 Ida Lupino-directed film `Hard, Fast and Beautiful'    \\
		\midrule
		\WR: &  \textcolor[rgb]{\colorone}{Sally Forrest} \textcolor[rgb]{\colortwo}{died on} \textcolor[rgb]{\colorthr}{March 15} at her home in Beverly Hills, California. \\
		\bottomrule
	\end{tabularx}
	\caption{A correct case from CNN/Daily Mail. The second half of the sentence is from the original text.}
	\label{tab:correct1}
	\end{subtable}
	
	\begin{subtable}{\linewidth}
    \centering
	\begin{tabularx}{\fulltablewidth}{p{\leftcolumnl} L}  
	    \toprule
        \Rel: &  Key relation: \{`subject': ` \textcolor[rgb]{\colorone}{Prince Harry}', `relation': ` \textcolor[rgb]{\colortwo}{is in}', `object': " \textcolor[rgb]{\colorthr}{attendance for England 's crunch match against France}"\}   \\
        \midrule
        \Gold: &  \textcolor[rgb]{\colorone}{Prince Harry} \textcolor[rgb]{\colortwo}{in} \textcolor[rgb]{\colorthr}{attendance for England's crunch match against France}. \\
        \midrule
        \NR: &  England beat France 55-35 in `Le Crunch'.\\
		\midrule
		\WR: &  \textcolor[rgb]{\colorone}{Prince Harry} \textcolor[rgb]{\colortwo}{in} \textcolor[rgb]{\colorthr}{attendance for England's crunch match against France}.\\
		\bottomrule
	\end{tabularx}
	\caption{A correct case from CNN/Daily Mail. Added relation covers through the sentence.}
	\label{tab:correct2}
	\end{subtable}
	
	\begin{subtable}{\linewidth}
    \centering
	\begin{tabularx}{\fulltablewidth}{p{\leftcolumnl} L}  
	    \toprule
        \Rel: &  Key relation: \{`subject': ` \textcolor[rgb]{\colorone}{valuable stock}', `relation': ` \textcolor[rgb]{\colortwo}{taken from}', `object': ` \textcolor[rgb]{\colorthr}{his antiques shop in Basingstoke}'\}  \\
        \midrule
        \Gold: &  Discovered \textcolor[rgb]{\colorone}{valuable stock} \textcolor[rgb]{\colortwo}{taken from} \textcolor[rgb]{\colorthr}{his antiques shop in Basingstoke}. \\
        \midrule
        \NR: &  Alan Stone, 51, arrested on suspicion of theft.    \\
		\midrule
		\WR: &  The father-of-four admitted he had a `lump in his throat' \\
		\bottomrule
	\end{tabularx}
	\caption{An incorrect case from CNN/Daily Mail. The generated summary does not follow the added relation at all.}
	\label{tab:incorrect1}
	\end{subtable}
	
	\begin{subtable}{\linewidth}
    \centering
	\begin{tabularx}{\fulltablewidth}{p{\leftcolumnl} L}  
	    \toprule
        \Rel: &  Key relation: \{`subject': ` \textcolor[rgb]{\colorone}{1,000 pieces}', `relation': ` \textcolor[rgb]{\colortwo}{is in}', `object': ` \textcolor[rgb]{\colorthr}{last two years}'\}  \\
        \midrule
        \Gold: &  Has inked \textcolor[rgb]{\colorone}{1,000 pieces} of art on leaves \textcolor[rgb]{\colortwo}{in} \textcolor[rgb]{\colorthr}{last two years}.\\
        \midrule
        \NR: &  Teacher Wang Lian has drawn hundreds of doodles on leaves for the last 10 years.    \\
		\midrule
		\WR: &  Teacher Wang Lian has drawn hundreds of doodles on leaves for the \textcolor[rgb]{\colorthr}{last two years}. \\
		\bottomrule
	\end{tabularx}
	\caption{An incorrect case from CNN/Daily Mail. Added relation wrongly affects the model and causes an error.}
	\label{tab:incorrect2}
	\end{subtable}
	
	\begin{subtable}{\linewidth} 
    \centering
	\begin{tabularx}{\fulltablewidth}{p{\leftcolumnl} L}  
	    \toprule
        \Rel: &   Key relation: \{`subject': ` \textcolor[rgb]{\colorone}{could first preventive tool}', `relation': ` \textcolor[rgb]{\colortwo}{is in}', `object': ` \textcolor[rgb]{\colorthr}{history}'\}  \\
        \midrule
        \Gold: &  WHO leader: This vaccine \textcolor[rgb]{\colorone}{could} be “the \textcolor[rgb]{\colorone}{first preventive tool} against Ebola \textcolor[rgb]{\colortwo}{in} \textcolor[rgb]{\colorthr}{history}"\\
        \midrule
        \NR: &  A vaccine will be tested \textcolor[rgb]{\colortwo}{in} a subsequent study.   \\
		\midrule
		\WR: &  The trial \textcolor[rgb]{\colorone}{could} be the \textcolor[rgb]{\colorone}{first preventive tool} \textcolor[rgb]{\colortwo}{in} \textcolor[rgb]{\colorthr}{history}. \\
		\bottomrule
	\end{tabularx}
	\caption{An incorrect case from CNN/Daily Mail. Error occurs beyond the added relation.}
	\label{tab:incorrect3}
	\end{subtable}
	
	\begin{subtable}{\linewidth} 
    \centering
	\begin{tabularx}{\fulltablewidth}{p{\leftcolumnl} L}  
	    \toprule
        \Rel: &   Key relation: \{`subject': ` \textcolor[rgb]{\colorone}{Former Premier League footballer Sam Sodje}', `relation': ` \textcolor[rgb]{\colortwo}{has appeared alongside}', `object': ` \textcolor[rgb]{\colorthr}{three brothers accused}'\}   \\
        \midrule
        \Gold: &  \textcolor[rgb]{\colorone}{Former Premier League footballer Sam Sodje} \textcolor[rgb]{\colortwo}{has appeared in court alongside} \textcolor[rgb]{\colorthr}{three brothers accused} of charity fraud.\\
        \midrule
        \NR: &  A \textcolor[rgb]{\colorone}{former} Leeds United defender has been charged with conspiracy to commit fraud.   \\
		\midrule
		\WR: &  \textcolor[rgb]{\colorone}{Former Premier League footballer Sam Sodje} \textcolor[rgb]{\colortwo}{has appeared alongside} \textcolor[rgb]{\colorthr}{three brothers accused} of fraud.  \\
		\bottomrule
	\end{tabularx}
	\caption{A correct case from XSum. The word `fraud' is inferred from the original text.}
	\label{tab:correct3}
	\end{subtable}
	\caption{Results Analysis}
	\label{tab:RA}
\end{table*}



\subsection{Case studies}
\cc{Through experiments, we found} that the summaries generated with added relations indeed keep the factual knowledge in the relation most of the time. However, there are still some cases where this solution can go wrong. We show some examples of correct and incorrect generation in Table~\ref{tab:RA}, and each of these examples is analyzed below.

\subsubsection{CNN/Daily Mail}
We first conduct experiments using CNN/Daily Mail. Here we only select the sentence that the relation best meets our requirements from the three to four sentences of a target, instead of using the whole target for training. By analyzing the first ten cases of CNN/Daily Mail and comparing the summaries generated with and without the added relation, we find that when a) the added relation covers thoroughly; b) there is a reference sentence in the original text, then the generated summary will be correct, as shown in Tables~\ref{tab:correct1} and~\ref{tab:correct2}.

However, the added relations only ensure that the generated summary is consistent with the facts described by the relation, and if the summary is not generated according to the added relations or the generated summary is beyond the coverage of the added relation, inconsistency may occur. This can be manifested in a) the generated summary does not refer to the added relation at all, as shown in Table~\ref{tab:incorrect1}; b) the generated summary extracts part of the content of the added relation, but this changes the correct summary, resulting in factual inconsistency, as shown in Table~\ref{tab:incorrect2}. c) the summary completely contains the keywords of the added relation, but factual inconsistency occurs beyond the added relation, as Table~\ref{tab:incorrect3} shows.

Since GPT-2 tends to extract the exact sentences from the original text when generating summaries for CNN/Daily Mail, the effect of this approach is likely to be amplified or obscured. To examine the performance of this approach on a more abstract dataset, we conduct the second experiment on XSum.

\subsubsection{XSum}
It is experimentally demonstrated that for a more abstract dataset like XSum, the generated summaries can be correct without referring to an exact sentence of the original text or beyond the coverage of added relations, as shown in Table~\ref{tab:correct3}. However, similar to CNN/Daily Mail, there is a possibility of error in the parts that the added relations do not cover, such as time, place, and so forth.

\subsection{Other discussions}

\subsubsection{CoCo's issue of measuring factual consistency}

{As we have seen in the experiments, even though $G{'}$ is more similar to the golden summary $R$, CoCo gives $G$ a higher score than $G{'}$. This is because} the summaries generated from the original texts $G$ are often more likely to have exact words from the original texts $X$ than the golden summaries $R$. This makes CoCo consider $G$ to be more factually consistent with $X$ than $R$, thus giving $G$ a higher CoCo score than $R$. And since our relations are drawn from $R$, most of the time the summaries generated on $X{'}$ are close to $R$, thereby making the CoCo score for $G{'}$ lower than $G$, despite $G{'}$ are often more consistent with the facts of the original text than $G$. This problem is more obvious on CNN/Daily Mail, as $G$ of CNN/Daily Mail include more exact sentences from $X$, while $G{'}$ are closer to the references $R$. For XSum, although $G$ and $G{'}$ are similarly abstract, the CoCo score for $G$ is still slightly higher than the CoCo score for $G{'}$. {An example from XSum is shown in Table~\ref{tab:CoCoscoringissue}}. Moreover, if we retain verbs in addition to nouns in the calculation of CoCo, the CoCo scores of $G$ and $G{'}$ will be further reduced. We speculate that this may be because, in the document summarisation task, the nouns in the original text are retained as is, but the verbs associated with a noun are often uncertain.

\begin{table*}[!htbp]
	\centering
	\setlength{\aboverulesep}{2pt}
    \setlength{\belowrulesep}{2pt}
        \centering
	\begin{tabularx}{\fulltablewidth}{p{\leftcolumnl} L}  
	    \toprule
        \Rel: & Key relation: {`subject': ` \textcolor[rgb]{\colorone}{Google}', `relation': ` \textcolor[rgb]{\colortwo}{has hired}', `object': “ \textcolor[rgb]{\colorthr}{creator of one web 's most notorious forums 4chan}"}    \\
        \midrule
        \Gold: &  \textcolor[rgb]{\colorone}{Google} \textcolor[rgb]{\colortwo}{has hired} the \textcolor[rgb]{\colorthr}{creator of one of the web's most notorious forums - 4chan}.\\
        \midrule
        \NR: & \textcolor[rgb]{\colorone}{Google} has appointed Chris Poole as its new administrator of \textcolor[rgb]{\colorthr}{4chan}. \\
		\midrule
		\WR: &  \textcolor[rgb]{\colorone}{Google} \textcolor[rgb]{\colortwo}{has hired} a \textcolor[rgb]{\colorthr}{creator of one} of the \textcolor[rgb]{\colorthr}{most notorious forums} on the internet. \\
		\bottomrule
	\end{tabularx}
    \caption{An example of CoCo scoring issue. Even though $G$ contains obvious errors, it scores higher than $R$ and $G{'}$.}
	\label{tab:CoCoscoringissue}
\end{table*}

\section{Conclusion}

This study explores the properties of prefix-tuning and proposes a knowledge-enhanced document summarisation approach that combines prefix-tuning and natural language prompts. Experiments show that although the effectiveness of this approach is hardly reflected by CoCo, it does improve the ROUGE scores of the generated summaries, and keep the summaries factually consistent with the added relations. In the future, the knowledge to be added to the text, the format of the natural language prompts to use, and the length of the prefixes to train could be directions worth further exploring. In addition, we hope that this approach can provide new ideas for other natural languages processing tasks such as Q\&A or information extraction.


\bibliographystyle{IEEEtran}
\bibliography{ref}

\end{document}